\pdfoutput=1
\documentclass[10pt,twocolumn,letterpaper]{article}

\usepackage{iccv}
\usepackage{times}
\usepackage{epsfig}
\usepackage{graphicx}
\usepackage{amsmath}
\usepackage{amssymb}

\usepackage{booktabs} 
\usepackage{multirow}
\usepackage{array}
\usepackage{footnote}
\newcommand{\myparagraph}[1]{\vspace{0.1em}\noindent\textbf{#1}}
\usepackage[linesnumbered,lined,vlined,ruled,commentsnumbered]{algorithm2e}

\def\eg{\emph{e.g}\onedot} 
\def\ie{\emph{i.e}\onedot}

\def\etal{\emph{et al}\onedot}

\usepackage[pagebackref=true,breaklinks=true,letterpaper=true,colorlinks,bookmarks=false]{hyperref}

\iccvfinalcopy 

\def\iccvPaperID{*} 

\usepackage{xpatch}
\newcounter{mybibstartvalue}
\xpatchcmd{\thebibliography}{%
  \usecounter{enumiv}%
}{%
  \usecounter{enumiv}%
  \setcounter{enumiv}{\value{mybibstartvalue}}%
}{}{}

\ificcvfinal\fi
\begin{document}

\title{A Novel BiLevel Paradigm for Image-to-Image Translation}
\author{Liqian Ma$^{1}$ \quad Qianru Sun$^{2,3}\thanks{Corresponding author}$
\quad Bernt Schiele$^{3}$ \quad Luc Van Gool$^{1,4}$\\
\\
\small
$^{1}$KU-Leuven/PSI, Toyota Motor Europe (TRACE)\quad  $^{2}$ National University of Singapore\\
\small
$^{3}$Max Planck Institute for Informatics, Saarland Informatics Campus \quad $^{4}$ETH Zurich \\ 
{
\small \texttt{\{liqian.ma,
luc.vangool\}@esat.kuleuven.be}} \\
{
\small
\texttt{\{qsun, schiele\}@mpi-inf.mpg.de}} 
}

\maketitle

\begin{abstract}
Image-to-image (I2I) translation is a pixel-level mapping that requires a large number of paired training data and often suffers from the problems of high diversity and strong category bias in image scenes. 
In order to tackle these problems,
we propose a novel BiLevel (BiL) learning paradigm that alternates the learning of two models, respectively at an instance-specific (IS) and a general-purpose (GP) level.
In each scene, the IS model learns to maintain the specific scene attributes.
It is initialized by the GP model that learns from all the scenes to obtain the generalizable translation knowledge.
This GP 
initialization gives the IS model an efficient starting point, thus enabling its fast adaptation to the new scene with scarce training data.
We conduct extensive I2I translation experiments on human face and street view datasets.
Quantitative results validate that our approach can significantly boost the performance of classical I2I translation models, such as PG$^2$~\cite{ma2017pose} and Pix2Pix~\cite{pix2pix}. 
Our visualization results show both higher image quality and more appropriate instance-specific details, e.g., the translated image of a person looks more like that person in terms of identity.

\end{abstract}
\section{Introduction}
\label{sec:intro}

Humans have the impressive ability to imagine new scenes from a few  descriptions or reference images. 
For example, given a single picture of a butterfly and a single  overview picture of a garden, we can easily imagine video sequences of this butterfly flying around in the garden.
This is, however, a very challenging image-to-image (I2I) translation task for machine learning models, as only a single training image is available. Some works call this one-shot image translation~\cite{Mehrotra2017, BenaimNIPS2018}.

\begin{figure}[t]
  \centering
  \includegraphics[width=1\linewidth]{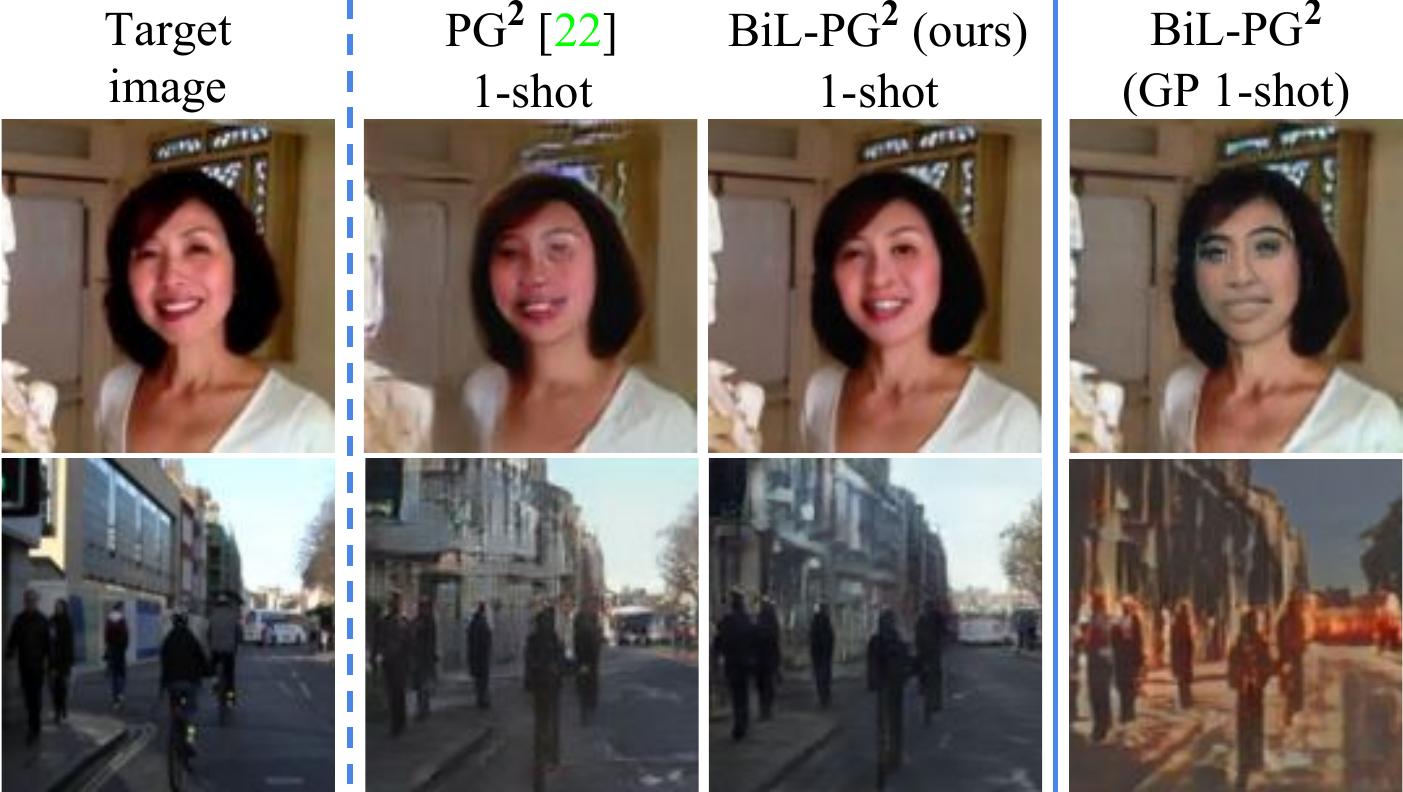}\\
     \caption{
     Our approach boosts the performance for the classical translation model PG$^2$~\cite{ma2017pose} for both face data and street view data. ``GP'' presents the generalizable translation knowledge we learned from diverse scenes. 
     }
    \vspace{-3mm}
  \label{teaser}
\end{figure}

By contrast, traditional I2I translation models~\cite{pix2pix,pix2pixHD,cycleGAN,ma2017pose} are usually trained on large-scale datasets of real-world images. 
These datasets, however, usually contain highly diverse image categories with only few samples in each category. 
It is difficult for the model to grasp the details of all categories. In other words, it is hard to learn a general I2I translation model.
For example in Fig.~\ref{teaser}, the pose guided person image generation model (PG$^2$)~\cite{ma2017pose}
suffers from artifacts, blurriness and the loss of individual characteristics in the translated images. 

In order to tackle these problems,
we propose a novel BiLevel (BiL) image translation
paradigm that alternates the learning between an instance-specific (IS) model and a general-purpose (GP) model.
The GP model aims to learn the generalizable translation knowledge across all image scenes, while each scene has an IS model that aims to maintain the specific attributes of the scene during translation.
In this manner, the parameters GP model can be quickly adapted to be an IS model each time when coming a new scene.
We achieve this by alternating the training processes of the IS and GP models.
Specifically, ``alternating'' means that when the training of the IS model finishes, the loss of a validation set (in the same scene) is used to optimize the GP model. 
In turn, the optimized GP parameters are used to initialize the IS model for the next scene. 
The validation loss is actually a generalization loss to optimize the GP model towards having a better generalization ability.

The GP model offers a warm start for the IS model to model a new scene and thus reduces the artifacts or blurriness caused by lacking data in the scene.
The IS model is fine-tuned from the GP parameters to the particular scene.
Therefore, it is more likely to generate instance-specific results.
As we can see from Fig.~\ref{teaser}, our approach successfully transfers the girl's identity features, such as skin color and hair style, to the output image and resolves the artifacts around the eyes, compared to PG$^2$~\cite{ma2017pose}.
Our result on the more challenging street view maintains a sharper and clearer appearance, e.g. for the glass windows.
%

At testing time, the IS model takes the GP model as point of departure and adapts to the new scene based on few training images.
An intuitive way to enhance IS training is to use these training images to query the available database for auxiliary data in similar scenes.
It is not obvious how to retrieve the most helpful data and how to utilize them, however, given that there are only a few images for each scene.
We propose a scene similarity metric that uses structure information and is computationally efficient.
We also propose to use the similar scenes to fine-tune the GP model - thus later providing
a better point of departure - rather than directly merge their data for IS training.
We call this method Auxiliary fine-tuning (called Aux in this paper).

\myparagraph{Our contribution} is thus three-fold.
(1) A novel and efficient BiLevel (BiL) approach that alternates the learning of GP and IS models.
It is a training paradigm orthogonal to the specific model architecture and can thus boost the performance of classic models.
(2) A simple and efficient Auxiliary fine-tuning (Aux) method that leverages data of similar scenes to enhance the few-shot learning in test scenes.
(3) Extensive experiments on two challenging I2I translation datasets -- FaceForensics~\cite{roessler2018faceforensics} and BDD100~\cite{yu2018bdd100k}.
Our results are based on two classic models, i.e. PG$^2$~\cite{ma2017pose} and Pix2pix~\cite{pix2pix}, and show that our approach can significantly boost performance both qualitatively and quantitatively.

\section{Related work}
\label{sec:related}
\myparagraph{Image-to-image translation.}
Image-to-image (I2I) translation aims to learn a mapping function to translate an image from a source domain to a target domain, \eg semantic maps to real images~\cite{pix2pix, pix2pixHD}, 
real images to cartoon images~\cite{taigman2016unsupervised}, gray-scale to color images~\cite{zhang2016colorful} and multi-domain translation~\cite{Zhao_2018_ECCV}.
Isola~\etal~\cite{pix2pix} proposed the Pix2pix method which solves the I2I problem with a conditional GAN~\cite{cGAN}. This currently is the most popular model. Pix2pixHD~\cite{pix2pixHD} is the high resolution version.
To alleviate the need for paired training data, Zhu~\etal proposed the self-supervising cycle-consistency loss~\cite{cycleGAN}. 
Both Pix2pix and CycleGAN learn deterministic mapping functions, \ie given one input, there is only one possible output.
To achieve multi-modal outputs, an implicit latent code is combined with the input~\cite{zhu2017toward,AugmentedCycleGAN}. 
Most recently, disentangled structure and appearance representations are explored to gain explicit control over the translation process. 
Some works use one image as structure reference (also called content) and another image as appearance reference (also called style or attribute) to generate a novel image~\cite{lee2018diverse,MUNIT,ma2018exemplar,joo2018generating,wang2018every}. 
Some other works propose to extract body landmarks from images as the structural reference~\cite{ma2017pose,ma2017disentangled,liu2018neural}.
Generally, the training of I2I models requires a large dataset that encourages the model to learn the structure of different objects.
Yet, this makes it hard to train a unified model as there are usually a handful of images for each object.
The intuitive way to address the problem is to fine-tune a large-scale trained model with few-shot data of a specific scene. 
This is not effective as the model tends to easily overfit to few samples.

\myparagraph{Few-shot I2I translation.}
Deep neural networks are data hungry, 
so moving to few-shot adaptation is an important improvement.  
Specific few-shot methods have been developed for different tasks such as recognition~\cite{vinyals2016matching,gidaris2018dynamic,qiao2017few,FinnAL17, RaviICLR2017, SunCVPR2019, LeeICML18, GrantICLR2018}, segmentation~\cite{rakelly2018few} and generative modeling~\cite{rezende2016one,edwards2016towards,hewitt2018variational}, but for 
I2I translation there is little work.
Benaim~\etal~\cite{benaim2018one} recently proposed a two-steps learning pipeline for one-shot unsupervised image translation.
Their assumption is that the source domain is one-shot but the target domain has abundant samples.
In contrast, our one-shot setting only contains a single pair of images, which is the most challenging case to study.
In terms of the optimization method, our alternating learning paradigm follows the same gradient descent method as~\cite{FinnAL17, RaviICLR2017, LeeICML18, GrantICLR2018, SunCVPR2019}.
These works focus on image classification and their base-learning task uses a small dataset randomly sampled from a large one. Our instance-specific learning focuses on specific objects, \eg a person identity and a street scene.

\myparagraph{Transfer learning}
Transfer learning transfers knowledge between related source and target domains~\cite{pan2010survey} and it is quite popular for addressing the small data problem~\cite{PanTKY11, wang2018transferring, AmirCVPR18}. The most intuitive and successful transfer fine-tunes a pre-trained model using the data of a new task~\cite{oquab2014learning}. 
Wang~\etal~\cite{wang2018transferring} evaluated how GANs can be transferred. They showed that using pre-trained networks boosts training and improves the quality of generated images.
We also initialize our bilevel training from a pre-trained network.

\begin{figure*}[h]
  \centering
  \includegraphics[width=1\linewidth]{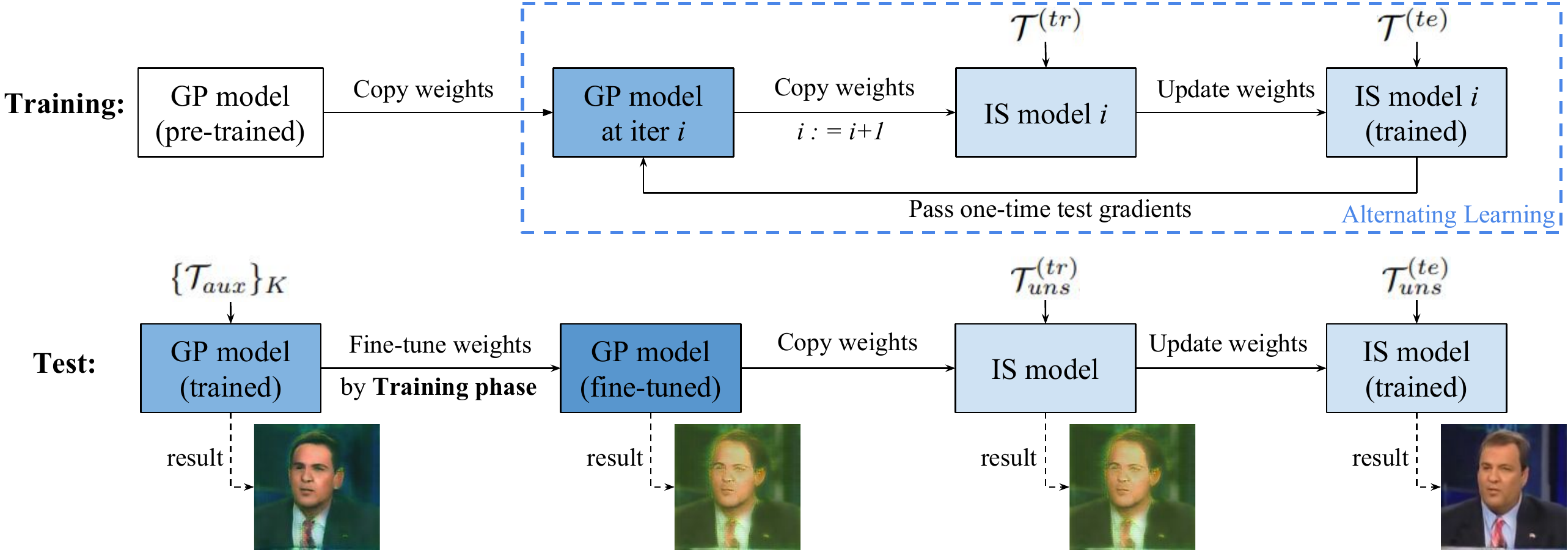}\\
     \caption{The proposed BiL image  translation paradigm pipeline. It contains two phases: training phase and test phase. During the training phase, the GP model is first initialized through pre-training. Afterwards, in each iteration of alternating learning, the IS and GP models are updated in turn. Specifically, the IS model initialized by the GP model is updated with episode training data $\mathcal{T}^{(tr)}$ and then used to calculate the test loss for GP model update. During GP-Testing, for each unseen task, the top $K$ most similar tasks $\{\mathcal{T}_{aux}\}_K$ are selected from the training set for fine-tuning the GP model according to which the new IS model is initialized and further updating.
     }
  \label{framework_overall_detail}
\end{figure*}

\myparagraph{Active learning.}
In our test phase, we select auxiliary data to fine-tuning the GP model, akin to active learning~\cite{settles2009active,vijayanarasimhan2010far,gavves2015active}.
The goal is to optimize the model within a limited time and annotation budget by selecting the most informative instances to annotate.
Vijayanarasimhan~\etal~\cite{vijayanarasimhan2010far} showed how to select a set of examples meeting a given budget of supervision.
Gavves~\etal~\cite{gavves2015active} proposed an active learning method to reuse existing knowledge (\ie available datasets).
This is close to our idea of reusing training data during testing.

\section{BiLevel I2I translation paradigm}
\label{sec:Meta_translation_def}

In this section, we first explain the problem setup for instance-specific image translation with few training images.
We then introduce our BiLevel (BiL) learning paradigm and the notations used for the two levels: general-purpose (GP) and instance-specific (IS).

\subsection{Problem setup}
\label{sec_translation_task}

The BiL learning paradigm contains two levels: the IS level for translating images in an individual scene; and the GP level for learning across 
scenes aiming to capture generalizable image translation knowledge.
We thus denote the space of all these scenes as $p(\mathcal{T})$ and each point $\mathcal{T}$ in $p(\mathcal{T})$ represents an I2I image translation task in an individual scene. 
For any $\mathcal{T}$, the goal is to translate the input image  to the target image that shares information with the input, e.g. the structure~\cite{pix2pix, pix2pixHD} or the object identity~\cite{ma2017pose, SunCVPR2018}.

This task specific definition follows the unified episodic formulation first proposed for few-shot image classification~\cite{VinyalsBLKW16}. 
Yet, there are several differences to the traditional paradigm: (1) the training of the GP model consists of a number of IS-level training-test procedures; (2) the training of an IS model is based on one episode with few-shot training image pairs and test image pairs; (3) the objective of the GP model is to initialize an IS model that should adapt to the new scene quickly; and (4) the final evaluation is the test result of the IS model adapted to unseen scenes.

\subsection{BiLevel training and test} 

The proposed BiLevel paradigm contains two phases: training and test, as shown in Fig.~\ref{framework_overall_detail}.
Note that the terms ``training'' and ``test'' are used only for the GP model. Because for each task, there are training and test procedures at the IS level.
We call them IS training and IS test for clarity.
There are multiple episodes $\{\mathcal{T}\}$ for training and one or multiple unseen scene episodes $\{\mathcal{T}_{uns}\}$ for test.
Each training episode $\mathcal{T}$ contains an IS training split $\mathcal{T}^{(tr)}$ as well as an IS test split $\mathcal{T}^{(te)}$.
The IS training loss on $\mathcal{T}^{(tr)}$ is used to optimize the IS model, while the IS test loss on $\mathcal{T}^{(te)}$ is used to optimize the GP model.
In the test phase, an IS model is initialized by the GP model to quickly adapt to the unseen scene through fine-tuning
on $\mathcal{T}^{(tr)}_{uns}$, then the IS test result on $\mathcal{T}^{(te)}_{uns}$ is reported as the final evaluation. If there are multiple unseen scene episodes, the average is reported.
Specific operations of BiL training and test are given as follows.

\myparagraph{Training phase.} 
Deep models trained from scratch usually converge slowly~\cite{pan2010survey}. Besides, it is notoriously difficult to train image generation models, e.g. Generative Adversarial Networks (GAN) based models, from scratch~\cite{ArjovskyB17}. 
Therefore, we first pre-train the GP model using the images from all training scenes, with the traditional training paradigm~\cite{ma2017pose, pix2pix}. The pre-trained weights are then simply taken as the initialization of the GP model which is used in the following training phase (see Fig.~\ref{framework_overall_detail}).
Afterwards, the GP model learns through a number of episodes in an alternating manner.
In each episode, it first initializes an IS model which is then adapted to a specific scene.
Finally, the validation loss computed from the learned IS model is used to update the GP model by meta gradient descent which unrolls the entire adaptation procedure of the IS model~\cite{FinnAL17}.

Given a training episode $\mathcal{T}$, the IS model parameters $\theta_{IS}$ are initialized by the parameters $\theta_{GP}$ of the current GP model which is learned by former tasks. Then, two stages, \ie episode training and episode test, will be executed.
In episode training, for each datapoint $x$ in $\mathcal{T}^{(tr)}$
the translation loss $L_{x\in \mathcal{T}^{(tr)}}(\theta_{IS})$
is used to optimize the IS model and the model parameters $\theta_{IS}$ become $\theta'_{IS}$ (see Eq.~\ref{eq:BiL_loss_IS}). 
After several epochs of learning on $\mathcal{T}^{(tr)}$, the updated $\theta_{IS}$ will be frozen for episode test.
During episode test, the losses of all datapoints $L_{\mathcal{T}^{(te)}}(\theta_{IS})$ are computed on $\mathcal{T}^{(te)}$, and the averaged loss is used to optimize $\theta_{GP}$ to be $\theta'_{GP}$ (see Eq.~\ref{eq:BiL_loss_GP}). This is done once in each episode.
\begin{align}
\theta'_{IS} &= \theta_{IS} - \alpha \nabla_{\theta_{IS}} L_{\mathcal{T}^{(tr)}}(\theta_{IS}) \label{eq:BiL_loss_IS},\\
\theta'_{GP} &= \theta_{GP} - \beta \sum\limits_{\mathcal{T} \sim p(\mathcal{T})}
    \nabla_{\theta'_{IS}} L_{\mathcal{T}^{(te)}}(\theta'_{IS}),
    \label{eq:BiL_loss_GP}
\end{align}
where $\alpha$ and $\beta$ are the step sizes for optimizing the IS and GP models, resp.

\myparagraph{Test phase.} 
In this phase, a new IS model for the unseen translation task $\mathcal{T}_{uns}$ will be trained with $\theta_{IS}$ initialized by the learned $\theta_{GP}$. 
Then, using the learned $\theta_{IS}$, the image translation results on $\mathcal{T}^{(te)}_{uns}$ provide the final evaluation.
Here, we propose an intuitive fine-tuning method to enhance $\theta_{GP}$ before IS training.
We leverage $\mathcal{T}^{(tr)}_{uns}$ as reference to quickly query auxiliary tasks $\{\mathcal{T}_{aux}\}_K$ (with similar scenes) from training data.
Then, we use the data to fine-tune $\theta_{GP}$.
We propose an efficient scene similarity metric using the structure information in images.
For example, the input of our model consists of image segmentation maps.
We calculate the similarity $Sim$ of two tasks by summing up the intersection-over-union (IoU) scores over all segmentation classes, as follows,
\begin{align}
    Sim(x_{uns},x_{aux}) = \sum_{c=1}^{C} \frac{\{x_{uns}=c\}\cap \{x_{aux}=c\}}{\{x_{uns}=c\}\cup \{x_{aux}=c\}}
    \label{eq:aux_task},
\end{align}
where $C$ is the number of object classes. $x_{uns}$ and $x_{aux}$ denote the input segmentation maps of the unseen task and a candidate task, resp.

After retrieval, we use the auxiliary data to fine-tune the GP model that later initializes a better starting point for the IS model (we call this method Aux.), rather than directly merge their data into the unseen task.
We will show the comparison in our experiments.


\section{BiLevel I2I translation models}
\label{sec:arch}
To evaluate the effectiveness of our BiLevel image translation paradigm, we choose two popular I2I translation problems: pose guided I2I translation~\cite{ma2017pose} and pix2pix I2I translation~\cite{pix2pix}. We test our approach with two classical models, \ie PG$^2$ model~\cite{ma2017pose} and Pix2pix model~\cite{pix2pix}, on mid-resolution images. 
Note that it is straightforward to apply our approach to higher resolution models, \eg Pix2pixHD~\cite{pix2pixHD}.

\subsection{Architectures}
\label{sec_gene_arch}
Our GP model and IS model share the same architecture which contains a generator $G$ and a discriminator $D$.
We apply the proposed BiLevel paradigm based on this architecture, and thus obtain four networks: GP-generator $G_{GP}$ and GP-discriminator $D_{GP}$; IS-generator $G_{IS}$ and IS-discriminator $D_{IS}$. 
Specifically, $G_{IS}$ and $D_{IS}$ work for the instance-specific translation and executes the translation task in a single scene. 
$G_{GP}$ and $D_{GP}$ provide a general-purpose basis as initialization for fast adaptation with only a few images in a new scene.

For the PG$^2$ model, 
the input of the generator is composed of a reference image and structure information. The reference image indicates the instance-specific content, \eg a street scene or a person identity.
The generator $G$ contains a residual U-net architecture consisting of an encoder and a decoder, both of which have several modules.
Each module consists of one down/up-sampling convolution block and one residual block following~\cite{ma2017pose}. 
As to the Pix2pix model, the architecture is similar except that the input only contains the structure information and the U-net architecture consists of convolution blocks instead of residual blocks, following~\cite{pix2pix}.

For both  PG$^2$  and Pix2pix models, the discriminator $D$ is a fully convolutional binary classifier whose input is either a fake generated image or a real ground truth image. It contains a series of down-sampling convolution blocks following~\cite{pix2pix}. Such fully convolutional discriminators can not only adapt to different image resolutions, but also penalize local structures.

\subsection{Losses}
\label{sec_loss}
We apply the L1 loss $L_1$ and the adversarial loss $L_{adv}$ to optimize the models as follows,
\begin{align}
    L_1(G) = &\mathbb{E}_{x_S,x_R,y}\big[\|G(x_S,x_R)-y\|_1\big],  
    \label{eq:L1_loss}  \\
    L_{adv}(G,D) = &{\mathbb{E}}_{x_S,y}\big[\log{D(x_S,y)}\big] \nonumber \\ 
    + &{\mathbb{E}}_{x_S,x_R}\big[\log{(1-D(G(x_S,x_R)))}\big], 
    \label{eq:GAN_loss}
\end{align}
where $x_S$ and $x_R$ denote the input structure information and reference image, resp., and $y$ denotes the target image. In the adversarial loss, $G$ tries to minimize this objective against an adversarial $D$ that tries to maximize it. 
In addition, we apply a recently proposed perceptual loss $L_P$, the Learned Perceptual Image Patch Similarity (LPIPS) metric~\cite{zhang2018unreasonable}, which is obtained by computing the $L2$ distance between the weighted deep features of images, as follows
\begin{align}
    L_P(G) = \mathbb{E}_{x_S,x_R,y}\big[\|\phi(G(x_S,x_R))-\phi(y)\|_2^2\big]
    \label{eq:LPIPS_loss},
\end{align}
where $\phi$ denotes the deep feature extractor which is an ImageNet pre-trained VGG16~\cite{VGG} in our experiments.
The full objective is thus as follows,
\begin{equation}\label{eq:total_loss}
    \min\limits_{G}\max\limits_{D} 
    L(G,D)=L_1(G)+\lambda_{a}L_{adv}(G,D)+\lambda_{b}L_{P}(G),
\end{equation}
where $\lambda_{a}$ and $\lambda_{b}$ denote weighting hyperparameters.

The learning objective is the same for the GP model and IS model. However, the optimization steps are different for them. Details are given in the following Sec.~\ref{sec:Learning}.

\subsection{Algorithms}
\label{sec:Learning}
Alg.~\ref{alg_train} and Alg.~\ref{alg_test} summarize the optimization procedures of training phase and test phase, resp.

\myparagraph{Training phase.} 
The details of the training phase are given in Alg.~\ref{alg_train}. 
To provide a good initialization for the GP model, we first pre-train it on a large-scale dataset (lines 1-5).
The pre-trained model is actually the conventional PG$^2$ or Pix2pix model, without BiLevel learning paradigm.
Then, we alternate the update of the GP model and the IS model (lines 6-22). 

\myparagraph{Test phase.}
The details of the test phase are given in Alg.~\ref{alg_test}. 
For each test task, we first select the auxiliary tasks and then fine-tune the GP model (lines 2-4).
We use the fine-tuned GP model parameters to initialize a new IS model (lines 5-7).
After the IS model has been adapted to the target task, we apply it to the generated images for the final evaluation (lines 8-13).

The optimization step sizes $\alpha$ and $\beta$ are adaptively controlled by the Adam~\cite{Adam} method. Basically, Adam~\cite{Adam} maintains a per-parameter learning rate adapted based on the average of the recent first and second moments of the gradients. This design enables fast and stable convergence of the optimization of both GP and IS models.

\begin{algorithm}
\caption{Training phase}
\label{alg_train}
\SetAlgoLined
\SetKwInput{KwData}{Input}
\SetKwInput{KwResult}{Output}
 \KwData{
 Translation training tasks 
 $\{\mathcal{T}\}$ and corresponding dataset $\mathcal{D}$; initial learning rate $\alpha$, $\beta$ for IS model and GP model\;}
 \KwResult{Generator $G_{GP}$, Discriminator $D_{GP}$}
 \%Pre-training \\
 Randomly initialize $G_{GP}$ and $D_{GP}$\;
 \For{samples in $\mathcal{D}$}{
 Update $G_{GP}$, $D_{GP}$ by Eq.~\ref{eq:total_loss} with step size $\alpha$\;
 }
 \%Alternating learning \\
 \For{meta-batches}{
 \%IS-update \\
 $G_{IS} \gets G_{GP}$\;
 $D_{IS} \gets D_{GP}$\;
 Pick up a task $\mathcal{T}$ from $\{\mathcal{T}\}$\;
  Sample training/test data $\mathcal{T}^{(tr)}$/$\mathcal{T}^{(te)}$ from $\mathcal{T}$ \;
 \For{samples in $\mathcal{T}^{(tr)}$}{
 Update $(G_{IS}$, $D_{IS})$ by  $L_{\mathcal{T}}(G_{IS}^\mathcal{T},D_{IS}^\mathcal{T};\mathcal{T}^{(tr)})$ in Eq.~\ref{eq:total_loss} with step size $\alpha$.\;
 }
  \%GP-update \\
  \For{samples in $\mathcal{T}^{(te)}$}{
 Compute $L_{\mathcal{T}}(G_{IS}^\mathcal{T}, D_{IS}^\mathcal{T};\mathcal{T}^{(te)})$ by Eq.~\ref{eq:total_loss}\;
 }
 $L_{GP} \gets Average(\{L_{\mathcal{T}}(G_{IS}^\mathcal{T}, D_{IS}^\mathcal{T};\mathcal{T}^{(te)})\})$\;
 Update $(G_{GP},D_{GP})$ by $L_{GP}$ with step size $\beta$\;
 }
\end{algorithm}

\begin{algorithm}
\caption{Test phase}
\label{alg_test}
\SetAlgoLined
\SetKwInput{KwData}{Input}
\SetKwInput{KwResult}{Output}
 \KwData{Translation train tasks $\{\mathcal{T}\}$, test tasks 
 $\{\mathcal{T}_{uns}\}$ and corresponding dataset $\mathcal{D}$, $\mathcal{D}_{uns}$; 
 initial learning rate $\alpha$, $\beta$ for IS model and GP model\;}
 \KwResult{Generated results $\hat{y}$}
 \For{each unseen task $\mathcal{T}_{uns}$}{
 \%GP-finetune \\
 Select $K$ auxiliary tasks $\{\mathcal{T}_{aux}\}_K$ from $\{\mathcal{T}\}$ by Eq.~\ref{eq:aux_task}\;
 Alternatingly update $(G_{GP},D_{GP})$ with $\{\mathcal{T}_{aux}\}_K$ by Alg.~\ref{alg_train} line 6-22\;
 \%IS-update \\
 $G_{IS} \gets G_{GP}$\;
 $D_{IS} \gets D_{GP}$\;
  Sample training/test data $\mathcal{T}_{uns}^{(tr)}$/$\mathcal{T}_{uns}^{(te)}$ from $\mathcal{T}_{uns}$ \;
     \For{samples in $\mathcal{T}_{uns}^{(tr)}$}{
     Update $(G_{IS}, D_{IS})$ by $L_{\mathcal{T}}(G_{IS}^\mathcal{T},D_{IS}^\mathcal{T};\mathcal{T}^{(tr)})$ in Eq.~\ref{eq:total_loss} with step size $\alpha$\;
     }
  \%Generate results \\
  $\hat{y} \gets G_{IS}(x_S,x_R)$
 }
\end{algorithm}
\section{Experiments}
\label{sec:exp}

We evaluate our BiLevel image translation paradigm in the instance-specific settings \footnote{In each task, an IS model initialized by the trained GP model gets a fast adaptation based on a few training images, then its result images are used for the evaluation.}.
We conduct extensive experiments for tackling two challenging problems: image-to-image translation (\ie Pix2pix model~\cite{pix2pix}) and structure guided image translation (\ie PG$^2$ model~\cite{ma2017pose}). 
We perform 1-shot and 5-shot I2I translation on face images, and 1-shot translation on street view images for which we conduct the model test in a challenging cross-dataset setting
\footnote{For more results, we refer the reader to the supplementary material.}. 

\subsection{Datasets}
\myparagraph{Face translation.} We use the FaceForensics~\cite{roessler2018faceforensics} dataset which contains $704$ training videos and $150$ test videos from news flashes by different reporters for face translation. 
We apply OpenPose~\cite{cao2017realtime} to detect facial landmarks for constructing the semantic segmentation maps with $7$ classes (regions): eyebrows, eyes, nose, lips, inner-mouth, face and background as shown in Fig.~\ref{fig:face_5_shot}. 
In addition, the exact face region is cropped according to the coordinate boundary of facial landmarks and then resized to $128\times 128$ pixels. 
To increase the number of training tasks, we divide each video in the training set into 7 short clips and apply different random image augmentations to 6 of them. For each video in the test set, we just use the original one.
After filtering out the facial landmark failures, we have 4,767 video clips for training and 148 for test, resp.

\myparagraph{Street view translation.} We use the training set of the Berkeley Deep Drive 100K (BDD100K) dataset~\cite{yu2018bdd100k} for training. BDD100K is
a real-world dataset with 7,000 segmentation-image training pairs
which are captured at diverse driving scenes and under various weather conditions. Since almost each segmentation-image pair comes from different video sequences captured at different scenes, we treat each pair as a single task for the 1-shot experiment. 
For testing, the performance is evaluated in a cross-dataset setting, where the model trained on the BDD dataset is tested on another dataset -- CamVid dataset~\cite{camvid}.  CamVid contains 704 annotated images with 33 semantic classes captured in 5 different street view videos. 
Specifically, we divide the CamVid videos into small video clips and each clip contains 10 successive annotated images. 
To compromise between the CamVid and BDD datasets, we map the 33 classes of CamVid to the 20 classes of BDD. Images are resized and then center-cropped to the size of 128$\times$256. 
For each video clip, we randomly select 1 image for 1-shot learning and 5 images for final evaluation.
%

\subsection{Metrics}
\label{sec:metrics}
We provide both qualitative and quantitative results. As to quantitative evaluation, three well-known objective image quality metrics are used: Structural Similarity (SSIM)~\cite{ImageQuality}, Mean Square Error (MSE), Peak Signal-to-Noise Ratio (PSNR), and one recently proposed perceptual distance: the Learned Perceptual Image Patch Similarity distance (LPIPS)~\cite{zhang2018unreasonable} which has been demonstrated to correlate well with human perceptual similarity~\cite{zhang2018unreasonable}.
The LPIPS is given by a weighted L2 distance between deep features of images, where we use AlexNet~\cite{Alex} pre-trained on ImageNet as feature extractor, similar to \cite{MUNIT}.
For SSIM and PSNR, higher scores are better. For LPIPS and MSE, lower scores are better.
We report the mean scores across $2960$ and $1380$ randomly selected pairs (20 pairs per task) for the test set for face and street view translation, resp. 

\subsection{Implementation and setting details}
For model optimization, we use the Adam~\cite{Adam} optimizer with $\beta_1=0.5$, $\beta_2=0.999$ and the initial learning rate of $0.0001$.
The mini-batch size for optimizing the general-purpose (GP) model is set to $5$ (\ie task number) and the mini-batch size for optimizing the instance-specific (IS) model is also set to $5$ (\ie image pair number). 
The GP model is optimized with $50k$ and $20k$ iterations during pre-training and meta-training, resp. 
The IS model is optimized with 20 iterations for each new task.
The loss weights are set to $\lambda_a=10$ and $\lambda_a=2$.

\myparagraph{Data augmentation.}
To encourage the network to handle large displacements, we perform data augmentation techniques: horizontal flip, crop, rotate. 
As to the PG$^2$ model, we only have one segmentation-image pair of a specific instance. 
Therefore, we manually generate the reference image from the ground truth image via data augmentation, \ie random horizontal flip\footnote{The horizontal flip is not applied to the reference image of face translation.}, crop and rotation, similar to the method used in~\cite{SwapNet}.

\myparagraph{Ablation settings.}\\
\textbf{PG$^2$:} the PG$^2$ model~\cite{ma2017pose} trained with training data.\\
\textbf{Pix2pix:} the Pix2pix model~\cite{pix2pix} trained with training data.\\
\textbf{PG$^2$ (n-shot):} the PG$^2$ model trained with training data, and then fine-tuned with n-shot target scene data during test phase.\\
\textbf{Pix2pix (n-shot):} the Pix2pix model trained with training data, and then fine-tuned with n-shot target scene data during test phase.\\
\textbf{BiL-X w/o Aux (n-shot):} the model X (\ie PG$^2$ or Pix2pix) trained by our BiLevel paradigm (without using auxiliary tasks) with training data, and then fine-tuned with n-shot target scene data during test phase.\\
\textbf{BiL-X (n-shot):} the model X (\ie PG$^2$ or Pix2pix) trained by our BiLevel paradigm with training data, and then fine-tuned with n-shot target scene data during test phase.\\
\textbf{BiL-X (GP n-shot):} the GP model of model X (\ie PG$^2$ or Pix2pix) trained by our BiLevel paradigm with training data.



\begin{figure*}[h]
  \centering
  \includegraphics[width=1\linewidth]{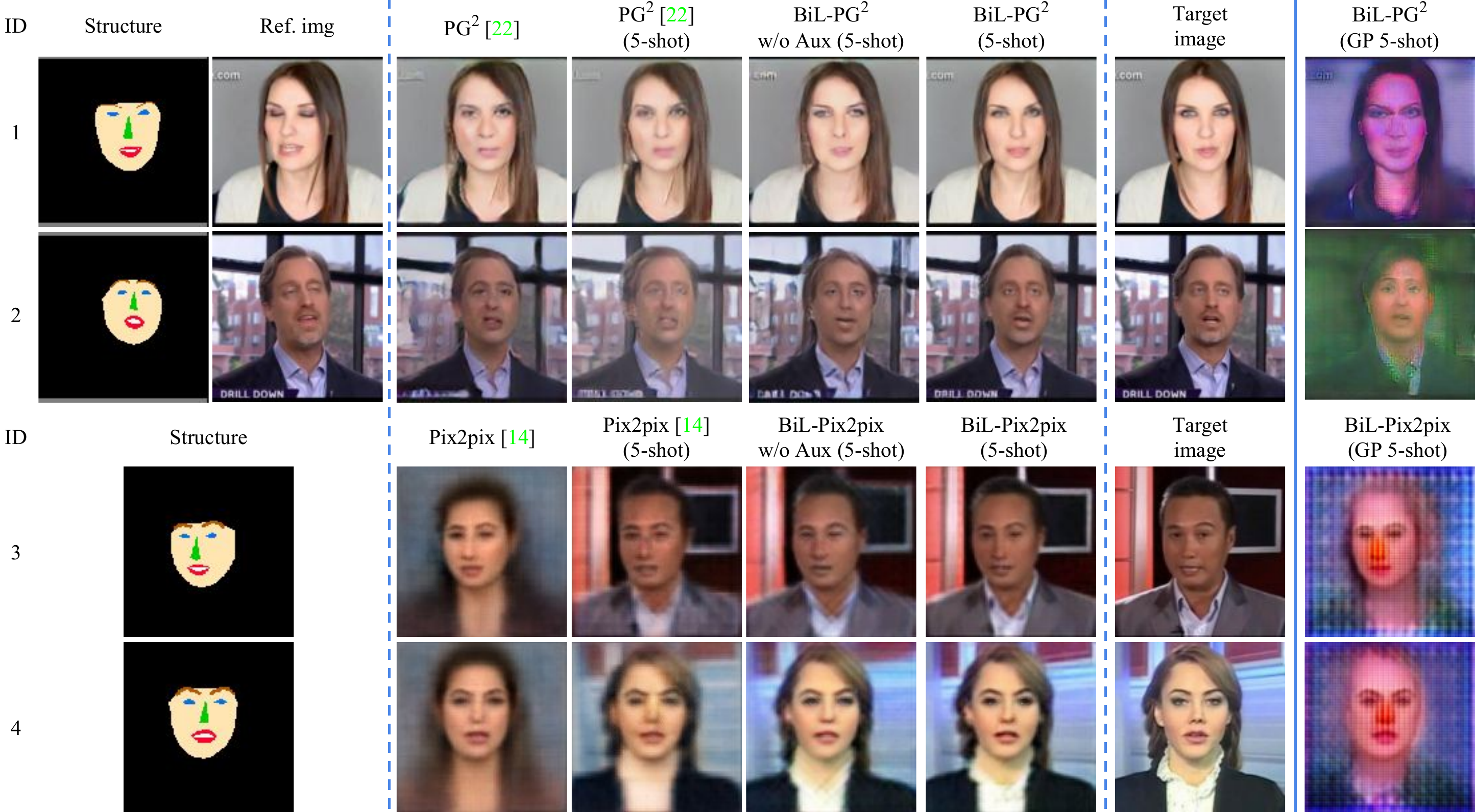}\\
     \caption{The 5-shot face image translation results on the FaceForensics dataset. From left to right: inputs, results of baselines and our methods, ground truth, and the output of GP model. Among the results, the second column denotes the baseline that is fine-tuned by 5-shot images of the target identity.}
  \label{fig:face_5_shot}
\end{figure*}

\begin{figure}[h]
  \centering
  \includegraphics[width=1\linewidth]{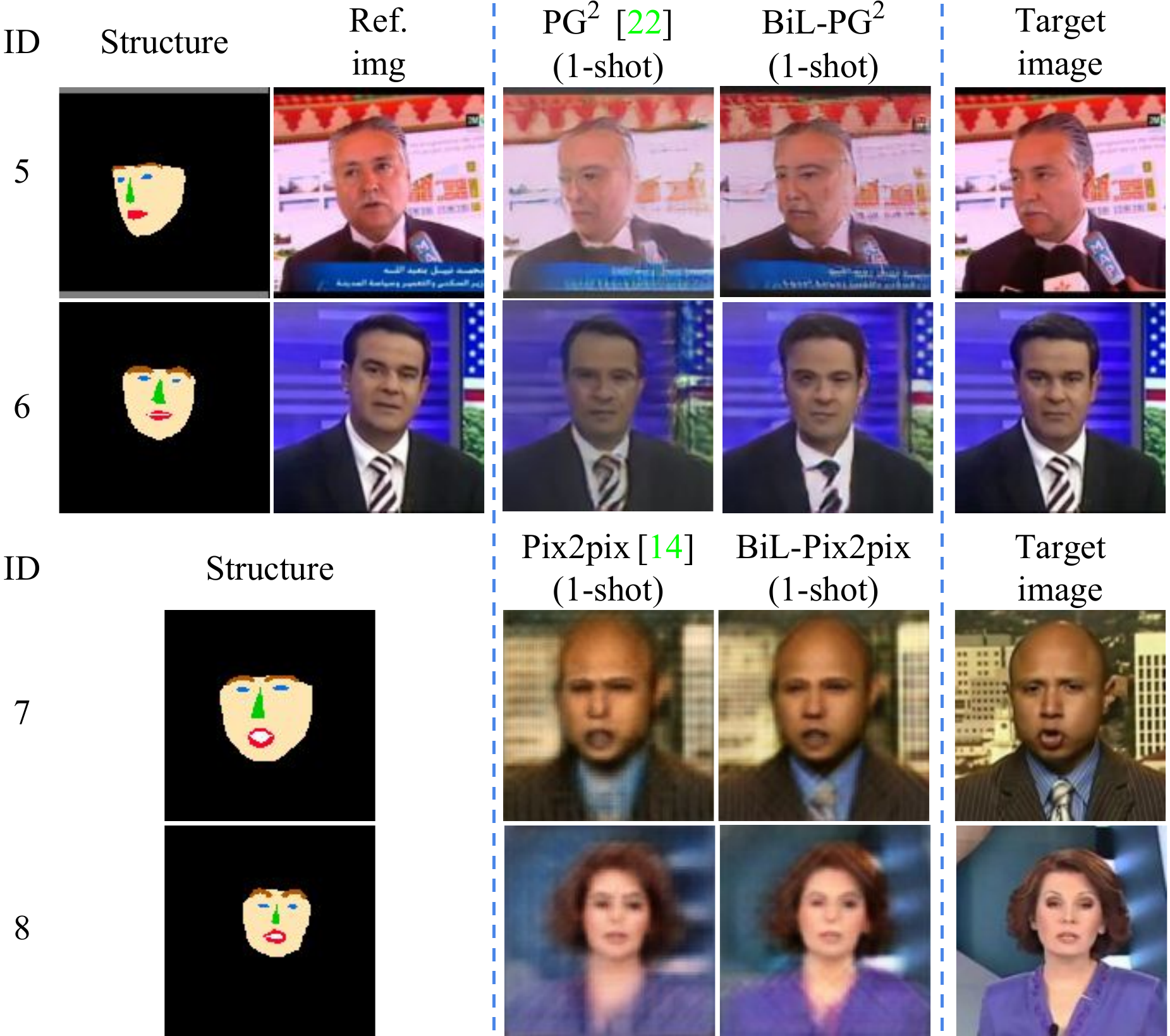}\\
     \caption{The 1-shot face image translation results on the FaceForensics dataset. From left to right: inputs, results of fine-tuned baseline and ours, and ground truth.}
  \label{fig:face_1_shot}
\end{figure}

\begin{figure*}[h]
  \centering
  \includegraphics[width=1\linewidth]{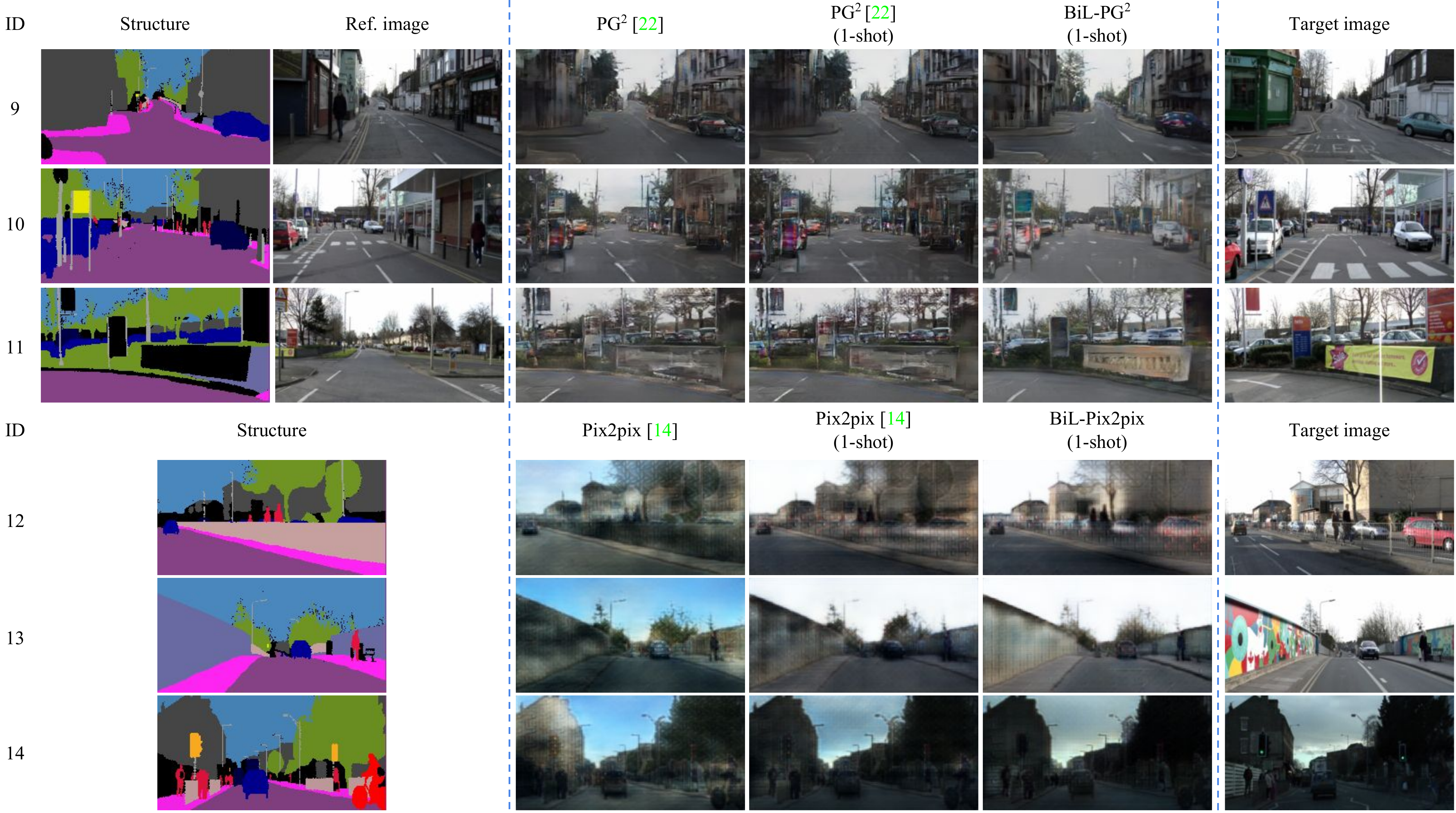}\\
     \caption{The 1-shot street-view image translation results in the challenging cross-dataset setting. From left to right: inputs, results of baselines and ours, and the ground truth.}
  \label{fig:street_cross}
\end{figure*}

\subsection{Face image translation}
\label{sec:exp_face}
For face translation, we evaluate our BiLevel approach in all ablative settings.
We use the 5-shot and 1-shot data on the FaceForensics dataset.

\myparagraph{Qualitative evaluation.}
As shown in Fig.~\ref{fig:face_5_shot}, the results generated by our BiL-PG$^2$(5-shot) model are more realistic and look quite closer to the target images. 
For example, comparing to the results of PG$^2$(5-shot), our results have the more and better-looking instance-specific details, such as the hair style and gender of ID-1, and the skin color, facial feature and age of ID-2.
Compared to the ablative model named BiL-PG$^2$ w/o Aux(5-shot), we have the conclusion that the auxiliary fine-tuning on the GP model greatly reduces the blurriness and artifacts.

As to Pix2pix model, the overall translation quality is inferior to PG$^2$ model due to the lack of reference image as input.
Relatively, our BiL-Pix2pix(5-shot) generates better results than the fine-tuned Pix2pix model (Pix2pix(5-shot)).
For example, the facial parts and hair styles of ID-3 and ID-4 have sharper appearances.

Additionally, Fig.~\ref{fig:face_1_shot} shows that our approach also shows reasonably good results in the most extreme 1-shot case. For example, BiL-PG$^2$(1-shot) successfully translates the poses of ID-5 and ID-6 It preserves better skin colors and more natural appearances of the faces, than the fine-tuned baseline, PG$^2$(1-shot).
Overall, the results above show that our BiL training approach brings impressive improvements for image-to-image translation in the extreme settings with scarce training data. 
Its results are not only with higher image quality but also maintaining more identity-specific details.


\myparagraph{Quantitative evaluation.}
The quantitative results are consistent with the qualitative visualizations. 
The 5-shot results are given in Tab.~\ref{tab:face_5_shot}.
We can see that our BiL-PG$^2$(5-shot) greatly reduce the LPIPS score from $0.076$ to $0.053$ ($30.2\%$) and increase SSIM score from $0.681$ to $0.756$ ($11.0\%$), comparing to the baseline model PG$^2$(5-shot).
Our BiL-Pix2pix(5-shot) also significantly reduces the LPIPS score by $40.4\%$ and increases the SSIM score by $19.4\%$, comparing to the baseline model Pix2pix(5-shot). 
In addition, the results of 1-shot setting are given in Tab.~\ref{tab:face_1_shot}.
As we mentioned, the LPIPS score is quite close to humans' evaluation. The large gains of LPIPS scores brought by our approach are greatly consistent with our improved visualization results in the figures.

\myparagraph{Analysis of the general translation knowledge.}
In the last column of Fig.~\ref{fig:face_5_shot}, we show the GP model results which demonstrate the visualization of the general knowledge encoded in our GP model, noting that the detailed appearance on the image is conditioned on the input image. We can see that this knowledge meets our expectation of an average-looking face with a random blurry background.

\begin{table}
\setlength{\tabcolsep}{5.5pt} 
\centering
\footnotesize
\begin{tabular*}{8.5cm}
{@{\extracolsep{\fill}} l c c c c }
\toprule 
Model & LPIPS$\downarrow$ & SSIM$\uparrow$ & MSE$\downarrow$ & PSNR$\uparrow$ \\
\midrule[0.6pt]	
	PG$^2$~\cite{ma2017pose} &  0.106 & 0.579 & 0.066 & 18.549  \\
	PG$^2$~\cite{ma2017pose} (5-shot) & 0.076 & 0.681 & 0.040 & 20.967  \\
    BiL-PG$^2$ (5-shot) w/o Aux & 0.054 & 0.755 & 0.028 & 22.987 \\
    BiL-PG$^2$ (5-shot) & \textbf{0.053} & \textbf{0.756} & \textbf{0.027} & \textbf{23.176}  \\
\midrule[0.6pt]	
	Pix2pix~\cite{pix2pix} & 0.357 & 0.357 & 0.271 & 12.106  \\
	Pix2pix~\cite{pix2pix} (5-shot) & 0.198 & 0.638 & 0.044 & 20.108  \\
    BiL-Pix2pix w/o Aux (5-shot) & 0.118 & 0.757 & 0.027 & 22.725 \\
    BiL-Pix2pix (5-shot) & \textbf{0.116} & \textbf{0.762} & \textbf{0.026} & \textbf{22.845} \\
\bottomrule[1pt]
\end{tabular*}
\caption{The evaluation scores for 5-shot face image translation on the FaceForensics dataset.} 
\label{tab:face_5_shot}
\end{table}

\begin{table}
\setlength{\tabcolsep}{5.5pt} 
\centering
\footnotesize
\begin{tabular*}{8.5cm}
{@{\extracolsep{\fill}} l c c c c }
\toprule 
Model & LPIPS$\downarrow$ & SSIM$\uparrow$ & MSE$\downarrow$ & PSNR$\uparrow$ \\
\midrule[0.6pt]	
	PG$^2$~\cite{ma2017pose} (1-shot) & 0.100 & 0.584 & 0.065 & 18.685  \\
    BiL-PG$^2$ (1-shot) w/o Aux & 0.085 & 0.630 & 0.050 & 19.887 \\
    BiL-PG$^2$ (1-shot) & \textbf{0.076} & \textbf{0.655} & \textbf{0.048} & \textbf{20.134} \\
\midrule[0.6pt]	
	Pix2pix~\cite{pix2pix} (1-shot) & 0.217 & 0.579 & 0.058 & 18.880  \\
    BiL-Pix2pix (1-shot) w/o Aux & 0.170 & 0.639 & 0.048 & 20.011 \\
    BiL-Pix2pix (1-shot) & \textbf{0.162} & \textbf{0.655} & \textbf{0.046} & \textbf{20.253} \\
\bottomrule[1pt]
\end{tabular*}
\caption{The evaluation scores for 1-shot face image translation on the FaceForensics dataset.}
\label{tab:face_1_shot}
\end{table}

\subsection{Street view image translation}
\label{sec:exp_street}

We evaluate our BiL paradigm on the more challenging street view image translation task, for which the images usually contain a large variety of objects.
We use a challenging cross-dataset setting: the GP model is trained on the BDD100K dataset then is tested on the CamVid dataset.
Due to the lack of annotated segmentation, we conduct only the 1-shot experiments.

\myparagraph{Qualitative results.} The 1-shot results are given in Fig.~\ref{fig:street_cross}.
Comparing with the fine-tuned baseline models PG$^2$(1-shot) and Pix2pix(1-shot), we can observe that our BiL models generate clearly more realistic images which contain a lot of street objects with various appearances.
For example, BiL-PG$^2$(1-shot) generates the trees with very similar appearances as in the target image (ID-9), buildings with sharp outlines (ID-10), cars with clear shapes (ID-11).

\myparagraph{Quantitative evaluation.}
The overall quantitative results of street view translation are provided in Tab.~\ref{tab:street_cross}.
It is easy to see that these are globally inferior to those of face translation (Tab.~\ref{tab:face_1_shot}). 
This reflects the greater challenge brought by the more complex data and more challenging cross-dataset setting. 
When comparing the results in Tab.~\ref{tab:street_cross}, we can conclude that our BiL models still generalize well in this harder setting, e.g. BiL-PG$^2$(1-shot) achieves a relative improvement rate of $8.0\%$ (from $0.174$ to $0.160$) for reducing LPIPS, and $11.3\%$ (from $0.400$ to $0.445$) for improving SSIM.

\begin{table}[t]
\setlength{\tabcolsep}{5.5pt} 
\centering
\footnotesize
\begin{tabular*}{8.5cm}
{@{\extracolsep{\fill}} l c c c c}
\toprule 
Model & LPIPS$\downarrow$ & SSIM$\uparrow$ & MSE$\downarrow$ & PSNR$\uparrow$ \\
\midrule[0.6pt]	
	PG$^2$~\cite{ma2017pose} & 0.194 & 0.373 & 0.060 & 15.652\\
	PG$^2$~\cite{ma2017pose} (1-shot) & 0.174 & 0.400 & 0.052 & 16.429 \\
    BiL-PG$^2$ (1-shot) w/o Aux & 0.166 & 0.434 & 0.045 & 16.933 \\
    BiL-PG$^2$ (1-shot) & \textbf{0.160} & \textbf{0.445} & \textbf{0.040} & \textbf{17.448} \\
\midrule[0.6pt]	
	Pix2pix~\cite{pix2pix} & 0.248 & 0.347 & 0.090 & 13.721  \\
	Pix2pix~\cite{pix2pix} (1-shot) & 0.203 & 0.459 & 0.040 & 17.883  \\
    BiL-Pix2pix (1-shot) w/o Aux & 0.184 & 0.507 & 0.028 & \textbf{19.327} \\
    BiL-Pix2pix (1-shot) & \textbf{0.184} & \textbf{0.509} & \textbf{0.028} & 19.322 \\
\bottomrule[1pt]
\end{tabular*}
\caption{The evaluation scores for 1-shot street-view image translation in the challenging cross-dataset setting.
} 
\label{tab:street_cross}
\end{table}
\section{Conclusion}
\label{sec:conclusion}

We introduced a BiLevel learning paradigm to learn general-purpose knowledge and fast model adaptability.
With the help of our BiLevel paradigm, the IS model of a new task can obtain the general-purpose experience from the GP model, and then quickly learn new instance-specific knowledge with only few-shot data.
Numerous quantitative and qualitative results demonstrate the effectiveness of our method. 
In addition, the images generated by our GP model can visualize
the general-purpose knowledge automatically.

\section*{Acknowledgments}
This research is in part funded by Toyota Motors Europe, the German Research Foundation (DFG CRC 1223), and NExT++ research supported by the National Research Foundation, Prime Minister's Office, Singapore, under its IRC@SG Funding Initiative.

{\small
\bibliographystyle{ieee}
\bibliography{egbib}
}


\end{document}